\documentclass{article}

\usepackage{PRIMEarxiv}

\usepackage[utf8]{inputenc} % allow utf-8 input
\usepackage[T1]{fontenc}    % use 8-bit T1 fonts
\usepackage{hyperref}       % hyperlinks
\usepackage{url}            % simple URL typesetting
\usepackage{booktabs}       % professional-quality tables
\usepackage{amsfonts}       % blackboard math symbols
\usepackage{nicefrac}       % compact symbols for 1/2, etc.
\usepackage{microtype}      % microtypography
\usepackage{lipsum}
\usepackage{fancyhdr}       % header
\usepackage{graphicx}       % graphics
\graphicspath{{media/}}     % organize your images and other figures under media/ folder
\usepackage[title]{appendix}
\title{ PDL: Regularizing Multiple Instance Learning with Progressive Dropout Layers
}

\author{
Wenhui Zhu\thanks{\textit{These authors contributed equally to this paper.}} \\
  School of Computing and Augmented Intelligence \\
   Arizona State University \\
  AZ 85281, USA\\
  \texttt{\ wzhu59@asu.edu} \\
   \And
  Peijie Qiu$^*$  \\
  McKeley School of Engineering \\
  Washington University in St. Louis \\
  St. Louis, MO 63130, USA\\
  \texttt{} \\
     \And
Xiwen Chen$^*$ \\
  School of Computing \\
  Clemson University \\
  Clemson, SC 29631, USA\\
  \texttt{} \\
     \And
  Oana M. Dumitrascu \\
  Department of Neurology \\
  Mayo Clinic \\
  Scottsdale, AZ 85251, USA\\
  \texttt{} \\
     \And
  Yalin Wang \\
  School of Computing and Augmented Intelligence \\
   Arizona State University \\
  AZ 85281, USA\\
  \texttt{} \\
}

\usepackage{algorithm}
\usepackage{algorithmic}
\usepackage{graphicx}
\usepackage{multirow}
\usepackage{makecell}
\usepackage{booktabs}
\usepackage{xcolor}

\usepackage{amsmath}
\usepackage{amsfonts}  
\usepackage{amsthm}
\theoremstyle{definition}
\newtheorem{definition}{Definition}
\begin{document}

\maketitle

\begin{abstract}
Multiple instance learning (MIL) was a weakly supervised learning approach that sought to assign binary class labels to collections of instances known as bags. However, due to their weak supervision nature, the MIL methods were susceptible to overfitting and required assistance in developing comprehensive representations of target instances. While regularization typically effectively combated overfitting, its integration with the MIL model has been frequently overlooked in prior studies. Meanwhile, current regularization methods for MIL have shown limitations in their capacity to uncover a diverse array of representations. In this study, we delve into the realm of regularization within the MIL model, presenting a novel approach in the form of a Progressive Dropout Layer (PDL). We aim to not only address overfitting but also empower the MIL model in uncovering intricate and impactful feature representations. The proposed method was orthogonal to existing MIL methods and could be easily integrated into them to boost performance. Our extensive evaluation across a range of MIL benchmark datasets demonstrated that the incorporation of the PDL into multiple MIL methods not only elevated their classification performance but also augmented their potential for weakly-supervised feature localizations. The codes are available at \href{https://github.com/ChongQingNoSubway/PDL}{https://github.com/ChongQingNoSubway/PDL}.

\end{abstract}

\section{Introduction}
Weakly-annotated data was a prevalent occurrence in numerous biomedical applications, including medical image segmentation~\cite{segmentation}, drug molecule discovery~\cite{drug}, and tumor detection~\cite{tumor}, etc. The weak supervisory signal comprised multiple instances (e.g., multiple tumor regions in a whole slide image) but was characterized by general categories (e.g., benign/malignant). Learning from this weakly-annotated data was typically formulated as the \emph{multiple instance learning} (MIL) problem. Instead of standard supervised learning such as image classification, the MIL assigned a label to a bag of instances rather than individually classifying each instance.

% \begin{figure}[htb]
% \centering
% \includegraphics[width=0.98\columnwidth]{IMG/loss_aaai_crop.png} % Reduce the figure size so it is slightly narrower than the column. Don't use precise values for figure width. This setup will avoid overfull boxes.
% \caption{The loss visualization compared existing MIL methods with PDL and without PDL based on Camlyon16 (complex dataset easily obervated overfitting). we observed that these existing methods still suffer from overfitting, even applied the DropInstance~\cite{li2021dual} and Dropout~\cite{shao2021transmil,zhang2022dtfd} at the beginning or end of MIL aggregator(eg., classifer layers). The proposed PDL was efficacy in addressing overfitting.  }
% \label{fig:loss}
% \end{figure}

Typical MIL frameworks consisted of an instance-level feature projector and a MIL aggregator. The feature projector mapped each instance onto a feature embedding, and then these embeddings were fed into the MIL aggregator for making predictions. Accordingly, prevailing MIL research focused on advancing either the feature projector~\cite{li2021dual} or the MIL aggregator~\cite{shao2021transmil,li2021dual,zhang2022dtfd,ilse2018attention,wang2018revisiting}. Despite the substantial efforts directed towards these two directions, MIL models continue grappling with the issue of \emph{overfitting} due to its weak supervisory signals. This enduring issue presented considerable challenges to their ability to attain intricate and expressive feature representations.

Regularization played a crucial role in addressing overfitting and improving the generalizability of neural networks~\cite{regularization_1,regularization_2,regularization_3}.  It motivated us to explore the values of adopting regularization in MIL models. Previous MIL models often integrated with two types of regularizations in their implementations: (i) randomly dropping instances (DropInstance) and (ii) adding Dropout~\cite{dropout_naive} to the last layer of the MIL model, as commonly applied in various image classification tasks~\cite{dropout,dropout_naive}.  
The DropInstance~\cite{li2021dual,shao2021transmil}, in which a specific proportion of instances was randomly dropped before entering the network (e.g., feature projector or  MIL aggregator), created various combinations of instances to serve as data augmentation. However, due to the limited number of positive instances (e.g., tumors), many combinations were similar, making it challenging to alleviate overfitting effectively.
Furthermore, mealy adding Dropout to the MIL aggregator classifier lacked in-depth modeling of instance correlations and proved insufficient in addressing overfitting. 
%the second method employed Dropout~\cite{dropout,dropout_naive} in the MIL aggregator~\cite{li2021dual,shao2021transmil}, which lacked in-depth modeling of instance correlations. Merely adding Dropout to the MIL aggregators classifier was also insufficient for addressing overfitting.
Besides, the regularization methods above were either applied at the beginning or the end of a MIL model. It cannot effectively assist the MIL model in discovering diverse and rich feature representations. In contrast, AttentionDropout~\cite{attdrop} dropped strongly activated elements in deep neural networks and enabled the network to learn more potential features.
It inspired us to explore a MIL model's regularization in the middle (i.e., the instance-level feature extractor). Nevertheless, the AttentionDropout lacked consideration of instance relations and localizations, and its main component was not directly applicable to MIL. Such challenges compelled us to explore a MIL-specific dropout method capable of mitigating overfitting while discovering latent features.

This paper proposed a novel progressive dropout layer (PDL) applied to the trainable middle layer within the instance projector. The PDL consisted of two main components: i) a MIL-specific attention-based dropout (MIL-AttentionDropout) and ii) a progressive learning scheduler. The MIL-AttentionDropout assigned a distinct drop rate to each instance based on its importance, enabling us to leverage the inter-instance correlations to discover richer and more representative features while introducing instance combination stochasticity to mitigate overfitting.  
The progressive learning scheduler adjusted the global maximum drop rates within MIL-AttentionDropout. We progressively increased the drop rates as the training progressed, effectively guiding the localizations within the MIL framework. 

The main contributions of this paper were as follows: 
\begin{itemize}
     \item We introduced an innovative MIL dropout method to combat overfitting, offering easy integrations into existing methods as a widely adopted paradigm.  
    \item We incorporated the capability to uncover latent features, specifically addressing defined challenges within the context of MIL, such as misleading localizations. It offered valuable insights for future investigations.
    \item Extensive experiments on various MIL benchmarks demonstrated the proposed method was effective in mitigating overfitting and discovering latent features to boost the performance of existing MIL methods in both classification and localization accuracy.  
\end{itemize}

\section{Related Works} 

\subsection{MIL methods:} 
The introduction of MI-Net~\cite{wang2018revisiting} has increased the prominence of bag-level MIL methods where only bag-level labels were used to supervise training. It mitigated the ambiguity of propagating bag-level labels to each instance in instance-level MIL methods~\cite{ feng2017deep, hou2016patch, xu2019camel}. Accordingly, 
empirical studies demonstrated that bag-level MIL methods generally exhibited superior performance when compared to their instance-level counterparts ~\cite{shao2021transmil, wang2018revisiting}. The mainstream focus of bag-level MIL methods mainly lied in advancing instance-level aggregation by incorporating attention mechanisms~\cite{ilse2018attention}, transformer~\cite{shao2021transmil}, knowledge distillation~\cite{zhang2022dtfd}, and non-local attention~\cite{li2021dual} to capture inter-instance correspondences within a bag. However, these methods introduced more complex model structures and enlarged parameter sets, increasing the risk of overfitting.
% In contrast, the proposed method mainly focused on mitigating the overfitting in MIL methods and thus orthogonal to existing MIL methods. 
Therefore, we proposed an approach orthogonal to existing MIL methods that could mitigate the overfitting and discover the latent features in MIL methods without altering the fundamental architectures of these methods.

\subsection{Dropout methods:}
Dropout has been experimentally validated as a potent technique for mitigating overfitting in diverse computer vision studies~\cite{dropout_naive, dropout}. Numerous investigations have ventured beyond the conventional dropout approach, exploring this avenue of research extensively. Examples included spatial-dropout~\cite{dropout}, dropout based on contiguous regions~\cite{ghiasi2018dropblock}, and AttentionDropout~\cite{attdrop}. In the context of MIL, where aggregators operated on instance embedding features without spatial context during training, the majority of studies used vanilla dropout.~\cite{dropout_naive, dropout}. This practice entailed the random deactivation of neurons within the bag-level classifier. Nonetheless, it neglected the intrinsic correlations existing within individual instances.
Unlike the standard Dropout method, AttentionDropout~\cite{attdrop} generated spatial attention maps via channel pooling and eliminated elements surpassing a prescribed attention weight threshold within a feature map. It encouraged the neural network to explore latent features more extensively. However, applying it directly to MIL faced challenges due to the lack of instance correlation modeling within instance-level embeddings and the use of hard thresholding, which overlooked inter-instance correlations.%However, direct application to MIL was unfeasible due to the absence of instance correlation modeling within instance-level embeddings, coupled with the issue of employing hard thresholding that disregarded inter-instance correlations. 
We presented an exhaustive analysis and discourse on these Dropout techniques in this study. Given the challenges encountered when adapting existing dropout methods to MIL, we introduced a novel MIL-specific dropout layer. This innovative approach capitalized on inter-instance correlations to dynamically modulate the dropout rates.
 
\begin{figure*}[t]
\centering
\includegraphics[width=1\textwidth]{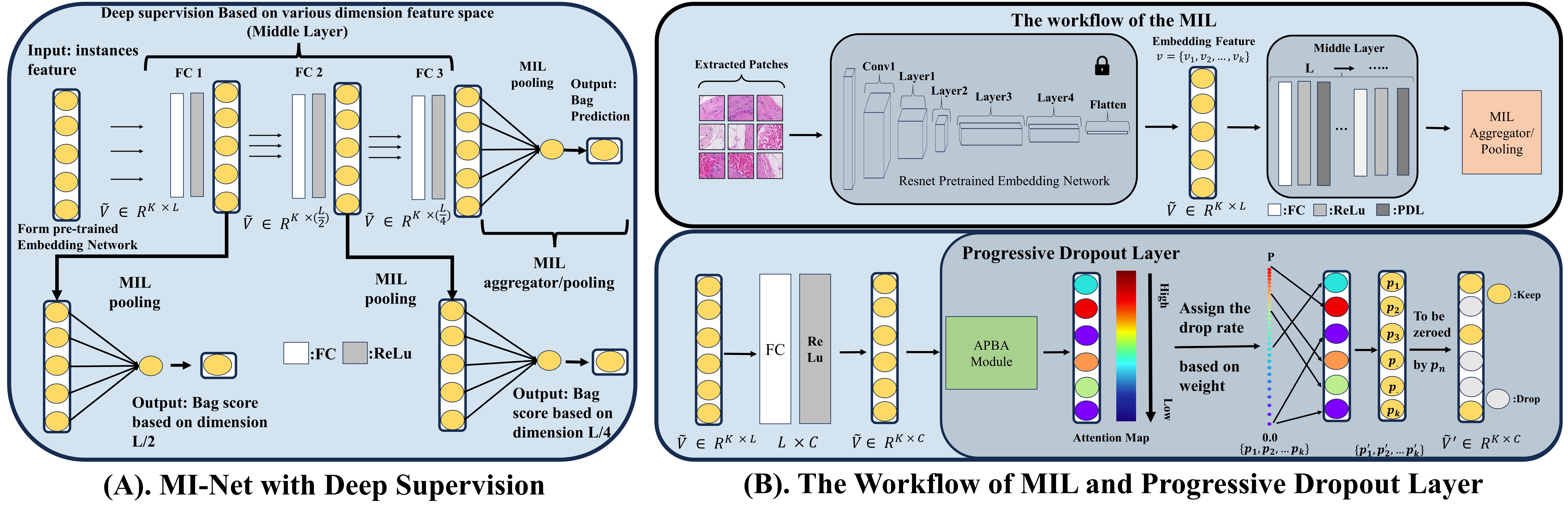} % Reduce the figure size to slightly narrower than the column.
\caption{(A). The MI-Net utilized the MIL pooling to enhance supervision across various embedding instance dimensions. (B). Illustration of progressive dropout layer (PDL) mechanism under the MIL workflow, taking one layer as an example. PDL dynamically assigned a drop rate to each instance based on Avarage-Pooling Based Attention (APBA), and applied Instance-Based Dropout to instances with drop rate $p'_{k}$. The Progressive Learning Scheduler controlled the global maximum drop rate $P$ to generate a set of $\left \{ p_{1}, p_{2}, \dots, p_{k} \right \}$. This mechanism enforced the model to discover latent features and mitigate overfitting.}
\label{workflow}
\end{figure*}

\section{MIL Preliminaries and Problem Statement}

\subsection{MIL Preliminaries}
 In the classical binary MIL classification, the objective is to learn a mapping from a given bag of instances $\textit{X}=\{x_k \ | \ k = 1, \cdots, K\}$ to a binary label $Y \in \{0, 1\}$. In most MIL applications, the instance-level labels $\{y_k  \ | \ k = 1, \cdots, K\}$ are unknown, making it a weakly-supervised problem:
\begin{equation}
    Y = \begin{cases} 0, \ \text{iff} \  \sum_{k} y_k = 0, \\ 1, \ \text{otherwise}. \end{cases}
    \label{eqn:1}
\end{equation}

We considered embedding-based MIL as an example, which typically consisted of two main modules: i) an instance-level projector and ii) a MIL aggregator. First, the instance-level projector $f_{\psi}(\cdot)$ parameterized by $\psi$ (e.g., a multi-layer perceptron) projected each instance $x_k$ within a bag into a feature embedding $\textbf{v}_{k} = f_{\psi}(x_k) $, with $\textbf{v}_k \in \mathbb{R}^{L}$, where $L$ denoted the embedding dimension. Secondly, the MIL aggregator was applied to combine instance embeddings into a bag prediction. The bag prediction was given as a Bernoulli distribution $P(X)$ (i.e., the probability of Y=1 given a bag $X$) by maximizing the log-likelihood.

\subsubsection{Instance-level projector.} 
The instance-level projector $f_{\psi}(\cdot)$ usually had two types of approaches. One approach was a pre-trained network (e.g., Resnet trained on Imagenet~\cite{shao2021transmil}, self-supervised manner~\cite{li2021dual}), which directly projected instances into embedding features and then fed them to the MIL aggregator for training. The other approach added the multilayer perceptron in the middle between the pre-trained network and the MIL aggregator. It could be used as feature dimensionality reduction~\cite{zhang2022dtfd} and provided deep supervisory across various embedding dimensions~\cite{wang2018revisiting} as shown in Figure~\ref{workflow}(A). It underwent joint training with the MIL aggregator. The second approach was referred to as the middle layer in subsequent discussions.

\subsubsection{MIL pooling.} 
The MIL aggregator was typically presented as a permutation-invariant function $\rho$, as implied by Eq.~\ref{eqn:1}. The pooling operation was a prevailing choice for such a permutation-invariant function, namely MIL pooling. The MIL could then be formulated as 
$    P(X) = g_{\theta}(\rho(\{ \textbf{v}_k\ | \ k=1, \cdots, K \}))$, 
where $g_{\theta}(\cdot)$ was a bag-level classifier parameterized by $\theta$.

We took attention-based pooling~\cite{ilse2018attention} as an example. Mathematically:
\begin{equation}
\begin{split}
     &\rho(\{ \textbf{v}_k\ | \ k=1, \cdots, K \}) = \sum_{k=1}^{K} \alpha_k \textbf{v}_k,  \\
     & \ \text{with} \ \ \alpha_k = \text{softmax}(\textbf{w}_1^T \text{tanh}(\textbf{w}_2 \textbf{v}_k^T)), 
\end{split}
\label{eq:abmil-pooling}
\end{equation}
where $\textbf{w}_1 \in \mathbb{R}^{D \times 1}$ and $\textbf{w}_2 \in \mathbb{R}^{D \times L}$ were learnable parameters. $\alpha_k$ implied the importance of the $k$-th instance.

\subsection{Problem Statement}
% The overfitting in MIL has garnered attention and attributed it to the lack of mutual-instance relationships~\cite{overfitting1,overfitting2,overfitting3,zhang2022dtfd}, The DTFD as the state-of-art method among them, which introduces pseudo-bags to enhance innate characteristics. However, this category of MIL methods still suffers from overfitting. Furthermore, there is a common regularization method known as DropInstance~\cite{li2021dual,shao2021transmil}, in which a specific proportion of instances is randomly dropped before entering the network(e.g., MIL aggregator or MIL aggregator with middle layer), creating various combinations of instances to serve as data augmentation. However, due to the limited number of positive instances (e.g., tumors or interest patches in images), many combinations are very similar, making it challenging to alleviate overfitting effectively. Moreover, some existing methods have incorporated Dropout~\cite{dropout,dropout_naive} in the MIL aggregator~\cite{li2021dual,shao2021transmil} or middle layer~\cite{wang2018revisiting}. However, it needs more in-depth considering instance correlation. 
In the context of MIL, Dropout, following the original formulation~\cite{dropout_naive,dropout}, can be represented as such: 
\begin{equation}
       % v\in R^{ n \times L } \ \overset{\alpha }{\rightarrow} \ v^{'} \in R^{ \alpha n \times L }
       O = (I \circ \xi), \; \text{with} \; \xi_{i,j} \sim \mathcal{B}(p),
       % mask\in R^{ n \times 1 }   \sim \mathcal{B}(p_{n})
       \label{eq:dropout}
\end{equation}
% where $I$ is the $B \times L$ matrix of input features for the current minibatch $B$, and $O$ is the $B \times L$ output matrix for the current layer.
where $I$ and $O$ were the $B \times L$ matrix of input and output features for the current minibatch $B$ in a dropout layer.
The $\circ$ denoted the elementwise (Hadamard) product of the input matrix with a $B \times L$ matrix of independent noise variables $\xi$, from a Bernoulli distribution with probability $1 - p$, with $p$ as the dropout rate.
Considering the obtained embedding feature of a bag of instances $\textbf{V} = \left \{ \textbf{v}_{1}, \textbf{v}_{2}, \dots, \textbf{v}_{K} \right \}$  with $\textbf{V}\in \mathbb{R}^{ K \times L }$ from the projector, the dropout mask $\xi$ is applied to each position of instance feature space $L$. The main research subjects were the forms of integrated single instances and building relationships among instances. The dropout operation disrupted the integrity of instances to increase uncertainty relations between instances. 
% Merely adding Dropout into MIL aggregators $g_{\theta}(\cdot)$ was insufficient for solving overfitting~\cite{shao2021transmil,zhang2022dtfd}.
% and the different model architectures further compound the lack of flexibility in incorporating Dropout as a paradigm.
Adding more Dropouts~\cite{dropout_naive,dropout} into MIL aggregators $g_{\theta}(\cdot)$ was an intuitive approach for solving overfitting. However, the issue was exacerbated by various model architectures, which posed challenges in effectively integrating Dropout as a viable paradigm.
Inspired by MI-Net (DS)~\cite{wang2018revisiting}, we proposed to employ Dropout in the middle layers as shown in Figure~\ref{workflow}(A). This paradigm offered two advantages:  the flexibility to add Dropout without affecting MIL aggregators and the ability to obtain representations of embedding features in different dimensions. The deep supervision method also signified the feasibility of acquiring instance weights on various dimensions by applying the MIL pooling. AttentionDropout~\cite{attdrop} was a natural choice and it has been extended to different dimensional features, forcing the network to discover latent features by dropping strong activation elements. However, it fell short in addressing overfitting. Its inherent capability to unveil latent features faced challenges within the context of MIL. For instance, its attention mechanism was not directly applicable to MIL, it overlooked instance correlations as shown in Eq.~\ref{eq:dropout}, and its threshold-based dropout led to misleading MIL localizations.

\section{Progressive Dropout Layer} 
The number of dropout layers could be adjusted based on the quantity of fully connected layers within the middle layer. To illustrate the principle, we have chosen a single layer for explanation, as depicted in Figure~\ref{workflow}(B). The PDL mainly included MIL-specific Attention-based Dropout and Progressive learning scheduler. The synergy between these two components aimed to imbue the capacity to unveil latent features and tackle overfitting when the PDL is employed.
In pursuit of this objective, we introduced a series of innovative MIL-specific approaches designed to overcome the constraints posed by existing dropouts that are not well-suited for MIL scenarios. 

\subsection{MIL-Specific Attention-Based Dropout} 
Given a set of dimensionality-reduced embedding features $\Tilde{\textbf{V}} = \left \{ \Tilde{\textbf{v}}_{1}, \Tilde{\textbf{v}}_{2}, \dots, \Tilde{\textbf{v}}_{K} \right \}$ with $\Tilde{\textbf{V}} \in \mathbb{R}^{ K \times C }$, concatenated from $K$ instances feature $\Tilde{\textbf{v}}_{k}\in \mathbb{R}^{ C }$, we introduced three key components within MIL-specific Attention-based Dropout by the following workflow order in figure~\ref{workflow}(B).
\subsubsection{Avarage-pooling based attention.} Rather than applying spatial attention focusing on each position for the feature map (e.g., embedding feature $v$ included $K \times C$ positions), we required an instance-level attention map at the current embedding dimension $C$, which would be employed in subsequent instance-based dropout to establish intrinsic connections within instances. The Attention-based pooling in Eq.~\ref{eq:abmil-pooling} was a primary approach to obtain the instance attention map in MIL. However, since this method introduced additional parameters $\textbf{w}_1$ and $\textbf{w}_2$, it would exacerbate overfitting. To address this problem, we proposed an approximate attention method, Avarage-Pooling Based Attention (APBA). It could be formulated as 
\begin{gather} 
\begin{aligned} A = & \ \text{softmax}    \left( \{\text{pool}(\Tilde{\textbf{v}}_k) \ | \ k =1, \cdots, K \}\right) \quad  \\  & \text{with} \ \text{pool}(\Tilde{\textbf{v}}_{k}) = \frac{1}{C} \sum_{i=1}^{C} \Tilde{\textbf{v}}_{k}^{(i)}
\end{aligned} \label{eqt:AAN} \raisetag{20pt} 
\end{gather} which employed the average pooling in each instance at embedding dimension $C$ after passing the softmax activation function to obtain the corresponding weight of each instance. The critical positions of each instance were activated after passing through a ReLU activation, and the intensity of each position was directly correlated to its contribution toward determining the bag label. The APBA served as an aggregator method to summarize the activated positions. Specifically, instances with more activated positions were considered positive instances with higher attention weights. The APBA was a non-parameterized approximation method that could identify the required positive instances.

\subsubsection{Dynamic drop rate assignment to each instance.}
When the network was forced to drop instances with high attention weights, it guaranteed the ability to discover latent features. Randomly dropping elements to help the network see a different set of data combinations effectively introduced a form of data augmentation to prevent network units from co-adapting~\cite{dropout_naive}. Accordingly, keeping the stochasticity of drop rates for instances with low attention weight was crucial in addressing overfitting. Based on this observation, we considered adjusting the drop rate of each instance based on the attention map. Here we proposed a non-linear interpolation method to generate a set of drop rates for each instance $\left \{ p_{1}, p_{2}, \dots, p_{k} \right \}$. Mathematically, it was formulated as 
\begin{equation} P/E* \log_{G}(linspace(0, G^{E} - 1, K)+ 1). \label{eqt:interpolation} 
\end{equation} 
It generated the number of instances $K$ drop rate set from $0$ to $P$. The $P$ denoted the global maximum drop rate. The $linspace(min,max,num)$ was a linear interpolation function to return $num$ evenly spaced samples from interval $[min, max]$, $E$, $G$ was a hyperparameter to control the spacing of produced set. As shown in Figure~\ref{workflow}(B), this method generated the drop rate set  $\left \{ p_{1}, p_{2}, \dots, p_{k} \right \}$ towards spacing close to the end of the vector ($P$), which ensured that instances with high weights received a correspondingly high drop rate. In contrast, other instances received drop rates proportionate to their weight ranks to maintain stochasticity. Specifically, the drop rate was allocated based on the attention weights of instances, ranging from high to low, corresponding in $\left \{ p_{1}, p_{2}, \dots, p_{k} \right \}$ from $0$ to $P$ descending order. Here we assigned the drop rate $\left \{ p'_{1}, p'_{2}, \dots, p'_{k} \right \}$ of each instance  $\left \{ \Tilde{\textbf{v}}_{1}, \Tilde{\textbf{v}}_{2}, \dots, \Tilde{\textbf{v}}_{K} \right \} $ based on the attention map $A$.
\subsubsection{Instance-based dropout.}
As illustrated in Eq.~\ref{eq:dropout}, most dropout methods lacked the consideration of instance correlations. To preserve the integrity of instances and eliminate the uncertainty between instances, we proposed the Instance-Based Dropout (IBD), represented as
\begin{equation}
       % v\in R^{ n \times L } \ \overset{\alpha }{\rightarrow} \ v^{'} \in R^{ \alpha n \times L }
       \Tilde{\textbf{V}}^{'} = (\Tilde{\textbf{V}} \circ \xi), \; with \; \xi_{k} \sim \mathcal{B}(p'_{k})
       % mask\in R^{ n \times 1 }   \sim \mathcal{B}(p_{n})
       \label{eq:DBI}
\end{equation}
where $\Tilde{\textbf{V}}^{'}$ was the $K \times C$ matrix of $K$ instances features $\left \{ \Tilde{\textbf{v}}_{1}, \Tilde{\textbf{v}}_{2}, \dots, \Tilde{\textbf{v}}_{K} \right \} $ output matrix. The $\circ$ denoted the elementwise product of the input matrix with a $K \times 1$ matrix of independent noise variables $\xi$, from a Bernoulli distribution with probability $1 - p'_{k}$, with $p'_{k}$ the dropout rate from an assigned corresponding set of $\left \{ p'_{1}, p'_{2}, \dots, p'_{k} \right \}$. Although Dropout methods~\cite{dropout,attdrop} were applied to each point in the entire feature map, our method operated on each instance, where the entire instance features would be dropped out.
\begin{table*}[t]
  \caption{Performance comparison on the MIL benchmark dataset. Each experiment was performed five times with a 10-fold cross-validation. We reported the mean of the classification accuracy ($\pm$ the standard deviation of the mean). The integration of the PDL model always resulted in enhanced performance.}
  \centering
  \resizebox{0.99\textwidth}{!}{
  \begin{tabular}{l|ccccc}
    \toprule
    Methods     & MUSK1  & MUSK2 & FOX & TIGER & ELEPHANT \\
    \hline
    mi-Net~\cite{wang2018revisiting} & 0.889 $\pm$ 0.039  & 0.858 $\pm$ 0.049 & 0.613 $\pm$ 0.035 &  0.824 $\pm$ 0.034 &  0.858 $\pm$ 0.037   \\
    MI-Net~\cite{wang2018revisiting} &  0.887 $\pm$ 0.041 & 0.859 $\pm$ 0.046 & 0.622 $\pm$ 0.038 & 0.830 $\pm$ 0.032 & 0.862 $\pm$ 0.034       \\
    MI-Net with DS~\cite{wang2018revisiting} & 0.894 $\pm$ 0.042 & 0.874 $\pm$ 0.043 & 0.630 $\pm$ 0.037 & 0.845 $\pm$ 0.039 & 0.872 $\pm$ 0.032    \\
    MI-Net with RC~\cite{wang2018revisiting} & 0.898 $\pm$ 0.043 & 0.873 $\pm$ 0.044 & 0.619 $\pm$ 0.047 & 0.836 $\pm$ 0.037 & 0.857 $\pm$ 0.040    \\
    ABMIL~\cite{ilse2018attention} &  0.892 $\pm$ 0.040 & 0.858 $\pm$ 0.048 & 0.615 $\pm$ 0.043 & 0.839 $\pm$ 0.022 & 0.868 $\pm$ 0.022   \\
    ABMIL-Gated~\cite{ilse2018attention} & 0.900 $\pm$ 0.050 & 0.863 $\pm$ 0.042 & 0.603 $\pm$ 0.029 & 0.845 $\pm$ 0.018 & 0.857 $\pm$ 0.027   \\
    % rFF+pooling ~\cite{pal2022bag} & 0.887 $\pm$ 0.037  & 0.860 $\pm$ 0.038 & 0.611 $\pm$ 0.041 & 0.828 $\pm$ 0.021 & 0.875 $\pm$ 0.030 \\
    % rFF+pooling-GCN ~\cite{pal2022bag} & 0.899 $\pm$ 0.030  & 0.860 $\pm$ 0.041 & 0.629 $\pm$ 0.034 & 0.829 $\pm$ 0.022 & 0.875 $\pm$ 0.030 \\
    % B-rFF+pooling-GCN ~\cite{pal2022bag} & 0.899 $\pm$ 0.036  & 0.872 $\pm$ 0.026 & 0.639 $\pm$ 0.027 & 0.830 $\pm$ 0.021 & 0.842 $\pm$ 0.034 \\
    % GNN-MIL~\cite{tu2019multiple} & 0.917 $\pm$ 0.048 & 0.892 $\pm$ 0.011 & 0.679 $\pm$ 0.007 & 0.876 $\pm$ 0.015 & 0.903 $\pm$ 0.010   \\
    DP-MINN~\cite{yan2018deep}  & 0.907 $\pm$ 0.036 & 0.926 $\pm$ 0.043 & 0.655 $\pm$ 0.052 & 0.897 $\pm$ 0.028 & 0.894 $\pm$ 0.030  \\
    NLMIL~\cite{wang2018non} & 0.921 $\pm$ 0.017 & 0.910 $\pm$ 0.009 & 0.703 $\pm$ 0.035 & 0.857 $\pm$ 0.013 & 0.876 $\pm$ 0.011 \\
    ANLMIL~\cite{zhu2019asymmetric} & 0.912 $\pm$ 0.009 & 0.822 $\pm$ 0.084 & 0.643 $\pm$ 0.012 & 0.733 $\pm$ 0.068 & 0.883 $\pm$ 0.014 \\
    DSMIL~\cite{li2021dual} & 0.932 $\pm$ 0.023  & 0.930 $\pm$ 0.020 & 0.729 $\pm$ 0.018 & 0.869 $\pm$ 0.008 & 0.925 $\pm$ 0.007 \\
    \hline
    ABMIL + PDL & \underline{0.991 $\pm$ 0.027} & \underline{0.962 $\pm$ 0.066} & \textbf{0.828 $\pm$ 0.058} & \underline{0.940 $\pm$ 0.049}  & \underline{0.970 $\pm$ 0.032}   \\
    ABMIL-Gated  + PDL  & \textbf{{0.993 $\pm$ 0.019}} & \textbf{{0.968 $\pm$ 0.049}} & \underline{{0.820 $\pm$ 0.066}} & \textbf{{0.941 $\pm$ 0.046}}  & {0.967 $\pm$ 0.033}  \\
    DSMIL  + PDL  & {0.987 $\pm$ 0.031} & \underline{{0.962 $\pm$ 0.048}} & {0.809 $\pm$ 0.062} & {0.933 $\pm$ 0.048}  & \textbf{{0.971 $\pm$ 0.035}}   \\
    % ABMIL-Gated \textbf{w/} SC & \textbf{0.969 $\pm$ 0.004} & \textbf{0.960 $\pm$ 0.008} & \textbf{0.791 $\pm$ 0.007} & \textbf{0.948 $\pm$ 0.004}  & \textbf{0.956 $\pm$ 0.004}
    \bottomrule
  \end{tabular}
  }
  \label{tab:classic}
\end{table*}
\subsection{Progressive Learning Scheduler}
A fixed threshold (unchanged maximum drop rate $P$ in Eq.~\ref{eqt:interpolation} for each training epoch) would mislead the MIL localization, leading to the network recognizing negative instances as target instances (positive instances). 
 % Differing from other computer vision tasks, MIL solely relies on bag-level supervision, emphasizing the significance of guiding the model effectively to learn the correct information. 
 We considered the following scenario. Positive instances with high attention weights initially dropped out when employing a fixed threshold (drop rate), and the classification task was retained to train under the previous bag label. It led the network to identify negative instances as positive instances when lacking the prior knowledge for positive instance recognition (See evidence in Figure~\ref{fig:scheduler}). To address it, we proposed a progressive learning scheduler to guide the MIL-specific Attention-based Dropout. The idea was to suppress the dropout layer during the initial training phases, thereby giving the ability to identify positive instances. After that, the dropout was progressively activated. We implemented it by allocating the parameter $P$ in Eq.~\ref{eqt:interpolation} based on $T$ epochs.

\begin{definition}
For any function $t \rightarrow P(t) $, here $P(0) = 0$  and $ \lim_{t\to T} P(t) = P_{max}$ was called progressive function, which adjusted the $P$ in Eq.~\ref{eqt:interpolation} for each epoch $t$. 
\end{definition}
 \noindent The initial condition $P(0) = 0$ where drop rate suppression was performed, dropout was gradually introduced in a way that $P(t) \leq  P_{max}$ for any $t$, finally convergence $P(t) \rightarrow P_{max} $. The progressive progress could be formulated as
\begin{equation}
    P(t) = P_{max} - \partial (t)P_{max} 
    \label{eq:PP}
\end{equation} 
Here $\partial (t)$ was a monotonic decreasing function for $ 0\leq \partial(t) \leq 1$, all satisfying the above criteria, were employed as a progressive learning scheduler.  The $P(t)$ generated $\left \{ P_{1}, P_{2}, \dots, P_{t} \right \}$ from range of 0 to $P_{max}$, which corresponds to each epoch $t$. Each global drop rate $P_{t}$ would feed to dropout layers to generate the set of instance drop rate $\left \{ p_{1}, p_{2}, \dots, p_{k} \right \}$ (Eq.~\ref{eqt:interpolation}, $P_{t}$ as $P$).

\begin{table*}[t]
\caption{Comparison results before and after adding the PDL to some state-of-the-art methods on the Camelyon16 and TCGA-NSCLC datasets. The classification accuracy (\%) and AUC (\%) were reported ($\pm$ the standard deviation of the mean) based on running five times of each experiment. $\Delta$ denoted the improvement gains after the integration of the PDL.}
\centering
     \resizebox{1\linewidth}{!}{
\begin {tabular}{p{4cm}ccccc|cccc}
\toprule %
 & & \multicolumn{4}{c}{Camelyon16} & \multicolumn{4}{c}{TCGA-NSCLC} \\
 \cmidrule(r){3-6} \cmidrule(r){7-10}
 & & \multicolumn{2}{c}{ImageNet Pretrained} & \multicolumn{2}{c}{ SimCLR Pretrained} & \multicolumn{2}{c}{ ImageNet Pretrained} & \multicolumn{2}{c}{ SimCLR Pretrained} \\
 \cmidrule(r){3-4} \cmidrule(r){5-6} \cmidrule(r){7-8}  \cmidrule(r){9-10}
 & & Accuracy & AUC & Accuracy & AUC & Accuracy & AUC & Accuracy & AUC \\
\toprule %

ABMIL~\cite{ilse2018attention} & \makecell{\\+PDL \\ $\Delta$} & \makecell{82.95 ± 0.51  \\ 83.98 ± 0.45 \\  \textbf{+1.03}} & \makecell{85.33 ± 0.31 \\ 85.61 ± 0.34 \\ \textbf{+0.28}} & \makecell{86.20 ± 0.34 \\ 88.84 ± 0.43 \\ \textbf{+2.64}} & \makecell{ 87.52 ± 0.75 \\ 91.09 ± 0.89 \\ \textbf{+3.57}}

& \makecell{88.29 ± 0.26 \\ 91.81 ± 0.52 \\  \textbf{+3.52}} & \makecell{94.39 ± 0.41 \\ 96.11 ± 0.42 \\ \textbf{+1.72}} & \makecell{89.61 ± 0.39 \\ 91.43 ± 0.33 \\ \textbf{+1.82}} & \makecell{ 95.20 ± 0.20 \\ 95.95  ±  0.12 \\ \textbf{+0.75}}

\\ \cmidrule (l ){1 -10}

ABMIL-Gated~\cite{ilse2018attention} & \makecell{\\+PDL \\ $\Delta$} & \makecell{82.64 ± 0.88 \\ 84.81 ± 0.69 \\  \textbf{+1.55}} & \makecell{85.22 ± 0.12 \\ 86.86 ± 0.69 \\ \textbf{+1.64}} & \makecell{86.31 ±  1.19\\  88.68 ± 0.43 \\ \textbf{+2.37}} & \makecell{88.39 ± 0.76 \\ 91.25 ± 0.46 \\ \textbf{+2.86}}

& \makecell{88.76 ± 0.63 \\ 92.19 ± 0.43 \\  \textbf{+3.43}} & \makecell{94.48 ± 0.54 \\ 96.20 ± 0.36 \\ \textbf{+1.72}} & \makecell{89.52 ± 0.33 \\ 90.67 ± 0.26  \\ \textbf{+1.15}} & \makecell{ 95.01 ± 0.14 \\ 96.03 ± 0.01 \\ \textbf{+1.02}}

\\ \cmidrule (l ){1 -10}

 DSMIL~\cite{li2021dual} & \makecell{\\+PDL \\ $\Delta$} & \makecell{80.46 ± 2.00 \\ 86.82 ± 1.64  \\ \textbf{+6.36}} & \makecell{82.46 ± 2.29 \\ 85.97 ± 0.67 \\ \textbf{+3.51}} & \makecell{84.34 ± 1.49 \\ 85.89 ± 0.85\\ \textbf{+1.55}} & \makecell{85.49 ± 2.77 \\ 88.85 ± 1.47 \\ \textbf{+3.33}}
 
 & \makecell{86.10 ± 0.62 \\ 89.52 ± 0.75 \\  \textbf{+3.42}} & \makecell{93.38 ± 0.77 \\ 94.15 ± 0.16 \\ \textbf{+0.77}} & \makecell{85.90 ± 0.72 \\ 89.05 ± 0.47 \\ \textbf{+3.15}} & \makecell{ 92.10 ± 0.68 \\ 94.27 ± 0.37 \\ \textbf{+2.17}}
 \\ \cmidrule (l ){1 -10}

 TransMIL~\cite{shao2021transmil} & \makecell{\\+PDL \\ $\Delta$} &  \makecell{80.62 ± 2.05 \\ 83.14 ± 0.74  \\ \textbf{+2.52}}  &  \makecell{81.48 ± 2.62 \\ 85.80 ± 0.68 \\ \textbf{+4.32}}  &  \makecell{87.44 ± 0.84 \\ 89.61 ± 0.69   \\ \textbf{+2.17}} &  \makecell{90.22 ± 1.17 \\ 92.01 ± 0.37 \\ \textbf{+1.79}}
 
 & \makecell{86.86 ± 0.72\\ 89.99 ± 0.72 \\  \textbf{+3.13}} & \makecell{93.16 ± 0.53 \\ 93.99 ± 0.18 \\ \textbf{+0.83}} & \makecell{88.73 ± 0.27  \\ 89.52 ± 0.48 \\ \textbf{+0.79}} & \makecell{ 94.87 ± 1.62 \\ 94.94 ± 0.28 \\ \textbf{+0.07}}
 
 \\ \cmidrule (l ){1 -10}

DTFD-MIL(MaxS)~\cite{zhang2022dtfd} & \makecell{\\+PDL \\ $\Delta$} & \makecell{83.33 ± 1.95 \\ 83.91 ± 1.60 \\ \textbf{+0.58}} & \makecell{86.07 ± 0.42 \\ 86.69 ± 0.28 \\ \textbf{+0.62}} & \makecell{85.74 ± 2.02 \\ 87.60 ± 0.78 \\ \textbf{+1.86}} & \makecell{90.72 ± 1.22 \\ 93.09 ± 0.80 \\ \textbf{+2.37}}

& \makecell{81.42 ± 0.58 \\ 85.05 ± 0.72 \\  \textbf{+3.63}} & \makecell{88.09 ± 0.22 \\ 90.67 ± 0.34 \\ \textbf{+2.58}} & \makecell{84.52 ± 0.47 \\ 86.82 ± 0.01 \\ \textbf{+2.30}} & \makecell{ 90.82 ± 1.00 \\ 91.49 ± 0.18 \\ \textbf{+0.67}}
\\
 
\bottomrule
\end{tabular}
}
\label{tab:wsi}
\end{table*}

\section{Experimental Results}
\subsection{Dataset and Experiments Detail} 
In this study, we conducted massive experiments to evaluate the performance of integrating the proposed PDL with several existing MIL methods. We used four public MIL benchmark datasets.
In this study, we extensively evaluated the performance of the proposed PDL integration with existing MIL methods using four public benchmark datasets.
\subsubsection{MIL benchmarks.}The benchmark datasets were MUSK1, MUSK2, FOX, TIGER, and ELEPHANT. These datasets were popularly studied to evaluate and compare the performance of MIL algorithms.
The MUSK1 and MUSK2~\cite{dietterich1997solving} focused on molecule classification. They contained a collection of bags, each consisting of instances representing atoms. Differently, FOX, TIGER, and ELEPHANT~\cite{andrews2002support} involve image classification. Each bag represented the images and contained instances that represented patches within images.

% analysis based on the different number of instances of each bag and the number of targets in each bag. 
\subsubsection{CAMELOYON16.} The primary objective of this dataset was to identify metastatic breast cancer in lymph node tissue. It consisted of high-resolution digital Whole Slide Images (WSIs). It was officially divided into a training set of 270 samples and a testing set of 129 samples. By the preprocessing approach detailed in~\cite{li2021dual}, each WSI was patched into non-overlapping patches with a size of $224 \times 224$. This procedure resulted in approximately 3.4 million patches when viewed at a magnification of $\times20$.

\subsubsection{TCGA-NSCLC.} The WSI dataset TCGA-NSCLC primarily identified two distinct subtypes of lung cancer: lung squamous cell carcinoma and lung adenocarcinoma. As outlined in~\cite{li2021dual}, 1037 WSIs were categorized into three sets: a training set encompassing 744 WSIs, a validation set comprising 83 WSIs, and a testing set containing 210 WSIs. After the preprocessing steps, approximately 10.3 million patches were extracted from these WSIs when viewed at a magnification level of $\times20$.

\subsubsection{MNIST-bags.}Following the dataset setting~\cite{ilse2018attention}, original MNIST dataset images were grouped into bags, each containing various digit images. The number of target images within a bag could vary; not all bags had the target digit. The digit 9 would be used as the target. The dataset was studied in the Discussion.

\subsubsection{Implementation details.} 
\label{sec:implementation}
We employed the ResNet~\cite{resnet} architecture as the pre-trained instance projector in WSI experiments to extract patch features. Two sets of pre-trained weights were utilized to ensure comprehensive evaluation, including pre-trained on the ImageNet and WSI patches following the contrastive learning framework called SimCLR~\cite{simplecontrastivelearning}. The SimCLR training settings were the same as DSMIL~\cite{li2021dual}. The feature of each patch is embedded in a 512-dimensional vector. The MIL benchmark already provided the pre-extracted embedding feature. For MNIST bags, we followed the ABMIL~\cite{ilse2018attention} and used the identical feature extractor method. All baselines were implemented with the parameter configurations specified in their original papers. To incorporate the PDL module, we employed a strategy involving adding three fully connected layers into the middle layer. A PDL layer was appended after each activation function, as illustrated in Figure~\ref{workflow}(B), the middle layer output dimension was 64, maximum drop probability $P_{max}$ = 0.45, and epoch $T = 200$ for WSI and  $T = 40$ for MIL benchmark. In this paper, all progressive learning schedulers adopted the non-linear interpolation function, same with Eq.~\ref{eqt:interpolation} ($K$ = $T$ and $P$ = $P_{max}$), which was satisfied with progressive progress in Eq.~\ref{eq:PP}, and all $G = 10$, $E = 0.5$ in this paper. All methods were implemented in PyTorch with NVIDIA A100. Furthermore, We reported other interpolation experimental results (Eqs.~\ref{eqt:interpolation} and~\ref{eq:PP}) and all detailed experiment settings in the supplementary material.

% \subsubsection{WSIs Instance Projector Pretrain Weight Details.} For two distinctly pre-trained ResNet-18 (i.e., ImageNet pre-trained and self-supervised pre-trained.  The self-supervised SimCLR manner employed the contrastive learning framework~\cite{simplecontrastivelearning} to pretrain the projector based on the training set, wherein contrastive loss training was implemented between extracted patches and corresponding two random data augmentation counterparts~\cite{li2021dual}. Pytorch officials provided the ImageNet pre-trained weight.  

\subsection{MIL Benchmark Results}
As shown in Table~\ref{tab:classic}, we integrated the PDL into  MIL aggregator methods, including ABMIL, ABMIL-Gated, and DSMIL. Following the setting in DSMIL, all experiments ran five times in the 10-fold cross-validation. The three aggregator methods with PDL outperforms all previous SOTA methods across all five MIL benchmark datasets. ABMIL + PDL achieved state-of-the-art accuracy by   82.8\% on FOX and improved accuracy by an average of 12.38\% on five datasets than previous ABMIL. ABMIL-Gated + PDL achieved state-of-the-art accuracy of 99.3\% on MUSK1, 96.8\%  on MUSK2, and 94.1\% on TIGER, and also improving accuracy by an average of 12.41\% on five datasets than previous ABMIL-Gated. DSMIL + PDL achieved state-of-the-art accuracy by 97.1\% on ELEPHANT, and PDL improved accuracy by an average of 5.54\% on five datasets.

\subsection{WSI Dataset Results}
We presented the results on two WSI benchmark datasets, Camelyon16 and TCGA-NSCLC. As shown in Table~\ref{tab:wsi}, a comparative study was performed to evaluate the performance gains by integrating the PDL. Four state-of-the-art aggregators were considered: ABMIL, DSMIL, TransMIL, and DTFD-MIL. To establish the authenticity of the experiments, each experiment was performed five times, and the mean and variance were computed. Through extensive experiments, we have observed that:  (1)  PDL improved all aggregators on both datasets. The average accuracy and AUC improvement of 2.26\%  and 2.43\% when applying the PDL on Camelyon16. It was also an average improved accuracy and AUC by 2.63\%  and 1.23\% on TCGA-NSCLC. In most cases, the variance was smaller than that of the original methods. Mitigating the overfitting made the model more stable. (2) The SimCLR pretraining feature extractor was better than pre-trained in ImageNet, which can be attributed to the fact that the SimCLR feature extractor has already learned the contextual information about WSI. In contrast, the PDL demonstrated a greater improvement on the feature extractor pre-trained in ImageNet, and the PDL could also improve performance even in the presence of noisy features. Remarkably, it achieved performance comparable to SimCLR in some cases, even when applied to features extracted from ImageNet pretraining. (3) The PDL in most cases, improved more on Camelyon16 than TCGA-NSCLC. The primary reason was that the TCGA-NSCLC dataset for detecting two types of lung cancer included a substantial portion of normal patches, which differed from conventional binary MIL problems, particularly when handling normal patches.

\begin{table}[t]
\centering
\caption{Comparison results between the PDL and other dropout methods on the CAMELYON16 dataset.}
\label{table2}
\resizebox{0.75\columnwidth}{!}{
\begin{tabular}{l|cc|cc}
\toprule %
     & \multicolumn{2}{c}{ABMIL} & \multicolumn{2}{c}{ABMIL-Gated} \\
    \cmidrule(r){2-3} \cmidrule(r){4-5}
     & Accuracy & AUC & Accuracy & AUC  \\ \cmidrule (l ){1 -5}
    Baseline & 86.82 & 86.38 &  85.27 & 88.19 \\ \cmidrule (l ){1 -5}
    Dropout & 87.60 & 87.61 & 86.05 &  89.57\\
    DropInstance & 86.05 & 88.90 & 87.60 & 88.97\\
    AttentionDropout & 88.37 & 85.35 & 85.27 & 86.13 \\
    SpatialDropout & 87.60 & 88.50 & 86.05 & 90.53\\
    PDL &  \textbf{89.15} &  \textbf{92.49}  & \textbf{89.15} &  \textbf{91.61} \\

\bottomrule
\end{tabular}
}
\end{table}

% \section{Analysis}

\section{Discussion}
\subsection{Comparison between the PDL and other dropout methods.}
We evaluated the PDL performance by comparing it with other Dropout methods, including Dropout~\cite{dropout}, SpatialDropout~\cite{dropout1d}, DropInstance~\cite{zhang2022dtfd} and AttentionDropout~\cite{attdrop}. All experiments performed extracted features from SimCLR pre-trained manner on the CAMELYON16 dataset, and ABMIL and ABMIL-Gated were used as baselines. All Dropout modules adopted the same method to embed into two aggregator methods. We applied a search for these methods within a drop rate range of 0 to 0.4, and the AttentionDropout used 0.65 as the threshold. Table~\ref{table2} shows the obtained optimal results. Our PDL outperformed other existing Dropout methods.

\subsection{Overfitting.}
\begin{figure}[t]
\centering
\includegraphics[width=0.5\columnwidth]{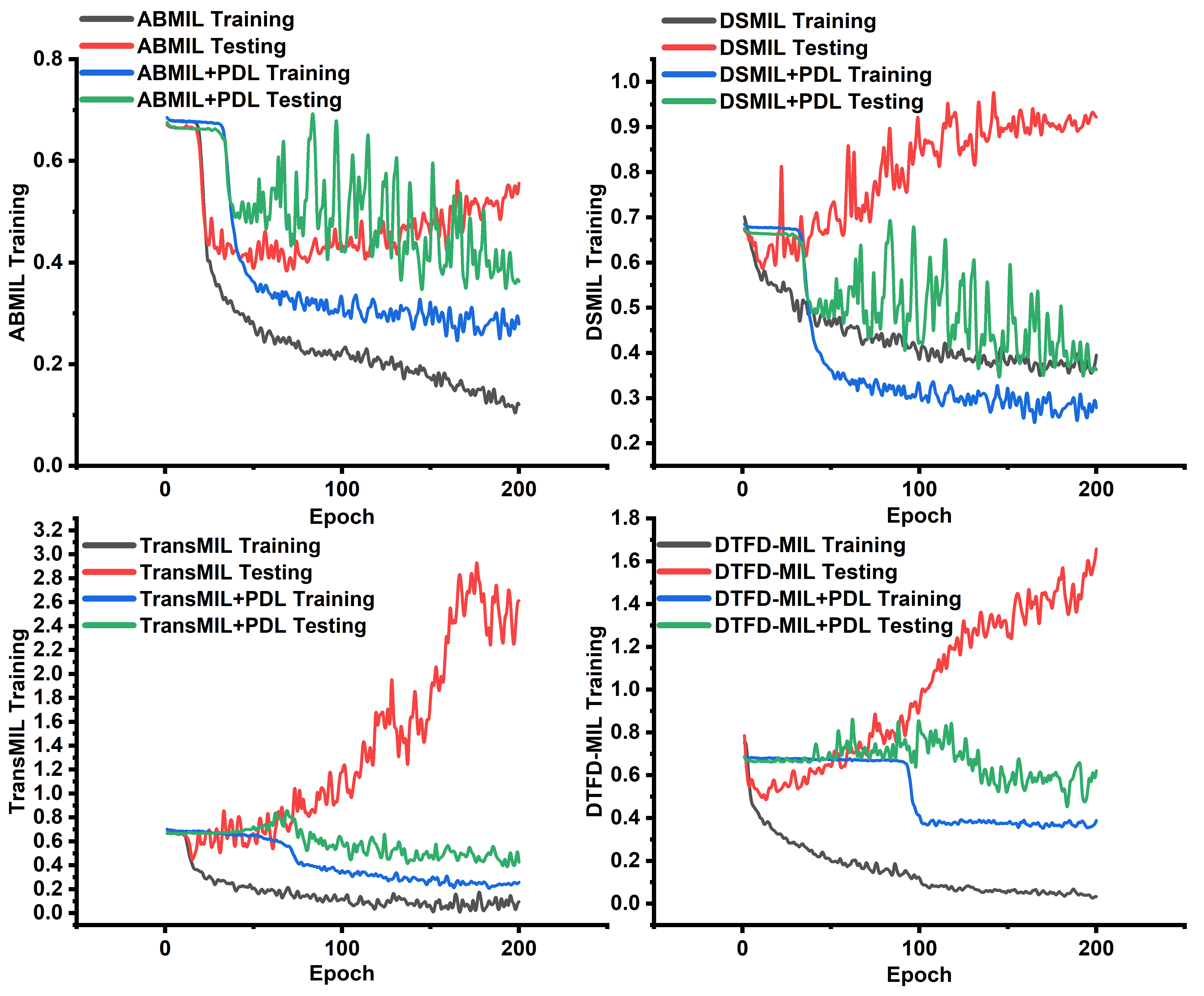} % Reduce the figure size so it is slightly narrower than the column. Don't use precise values for figure width. This setup will avoid overfull boxes.
\caption{The Camelyon16 experiment loss visualization before and after the integration of PDL during training.}
\label{fig:loss}
\end{figure}
As shown in Figure~\ref{fig:loss}, these MIL aggregator methods commonly exhibited overfitting, especially TransMIL and DTFD. The main reason was that these aggregator methods adopted more complex models and needed to learn a larger number of parameters. The proposed PDL demonstrated efficacy in addressing overfitting. All MIL aggregator methods integrated with PDL exhibited more stable declines in losses throughout the training process.

\subsection{WSI localization.}
\begin{figure}[t]
\centering
\includegraphics[width=0.98\columnwidth]{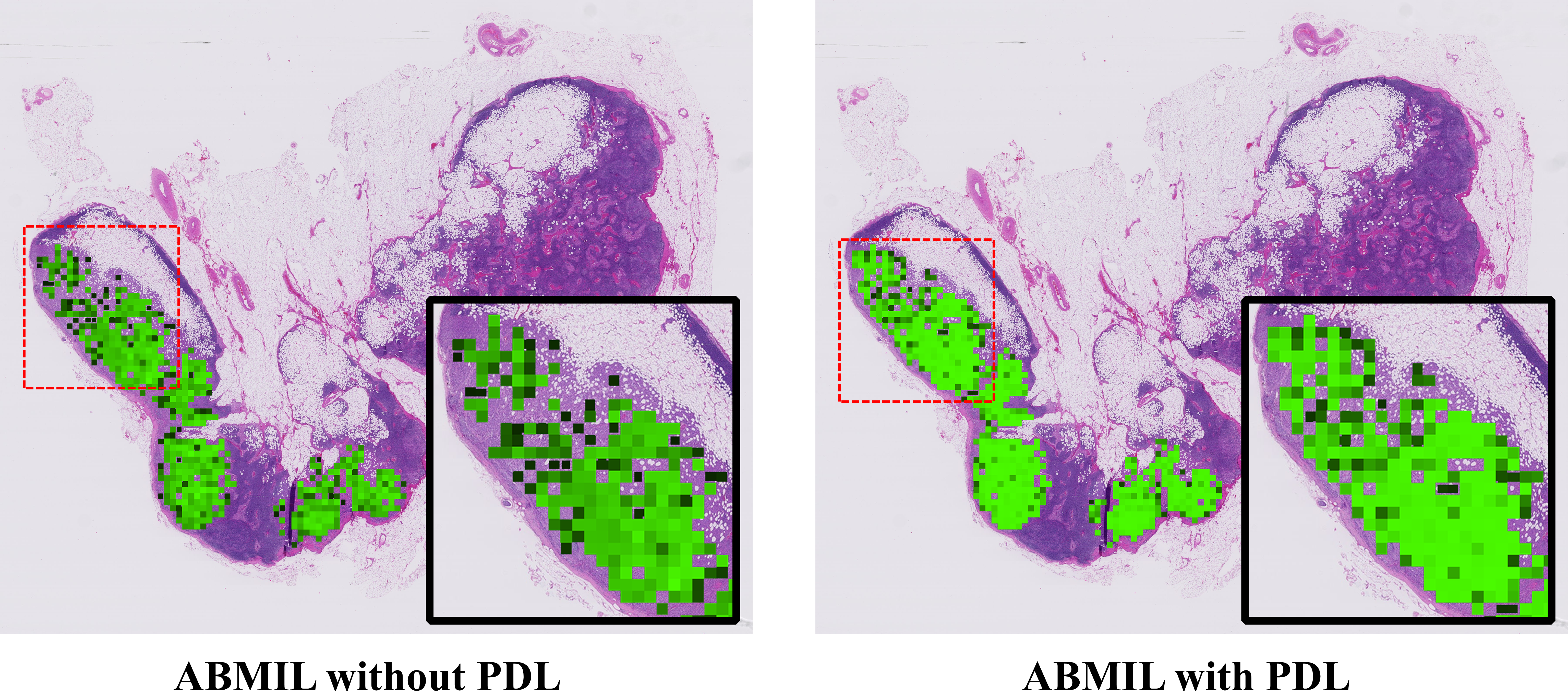} % Reduce .
\caption{Tumor localization in WSIs comparing ABMIL with PDL and Without PDL based on Camlyon16. }
\label{fig:localization}
\vspace{-0.2cm}
\end{figure}
As shown in Figure~\ref{fig:localization}, we visualized a tumor detection example based on the attention map of the ABMIL aggregator. The attention scores ranged from 0 to 1, where the brightness of green indicates the weight score, with brighter green representing higher scores. Compared to the ABMIL without PDL, the ABMIL integrating PDL provided more lesion patches and denser lesion localizations, with higher weight scores assigned to the lesion patches. The PDL not only effectively addressed overfitting but also showcased prowess in uncovering latent lesion features.

\subsection{Could PDL effectively assign drop rates as expected? }
\begin{figure}[t]
\centering
\includegraphics[width=1\columnwidth]{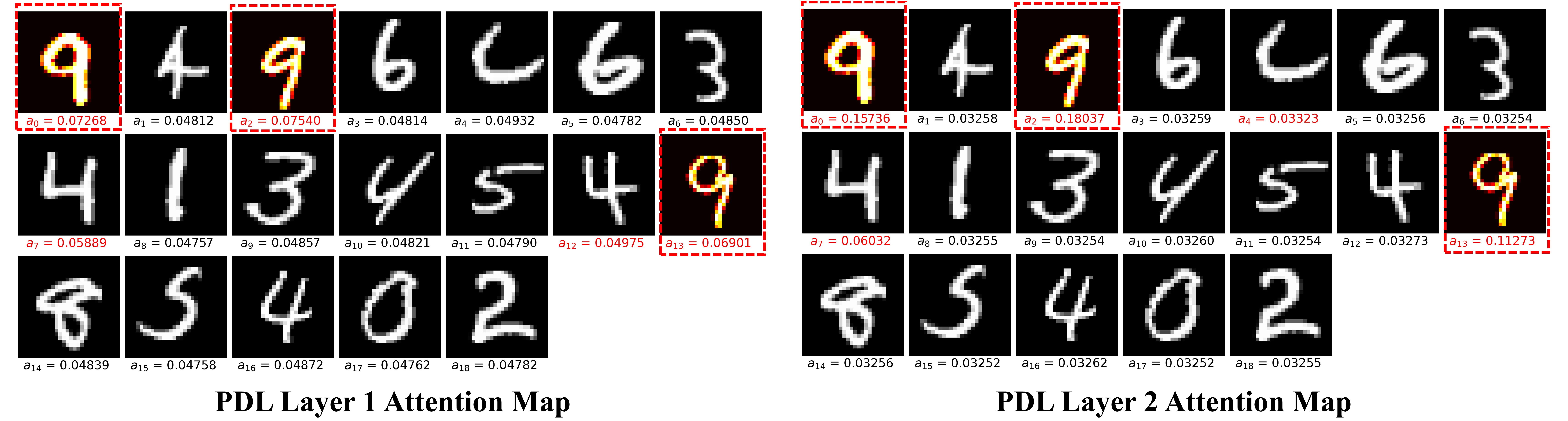} % Reduce the figure size so that it is slightly narrower than the column. Don't use precise values for figure width. This setup will avoid overfull boxes.
\caption{The APBA visualization of PDL in training, the number represents the attention weight, and the red block denotes the assigned top three drop rate.}
\label{fig:pdlatt}
\end{figure}
The APBA is a non-parametric attention method in PDL, which obtains the attention map in the given embedding dimension. We experimented on the MINIST dataset and validated whether PDL works correctly by visualizing the attention maps and drop masks during training. We conducted experiments applying the ABMIL aggregator method and two-layer PDL. The PDL can capture the target instances `9' at the different embedding features (see Figure~\ref{fig:pdlatt}), which gave the highest attention weight while being assigned the highest drop rate.

\subsection{Why was the progressive learning scheduler necessary?}
\begin{figure}[htb]
\centering
\includegraphics[width=1\columnwidth]{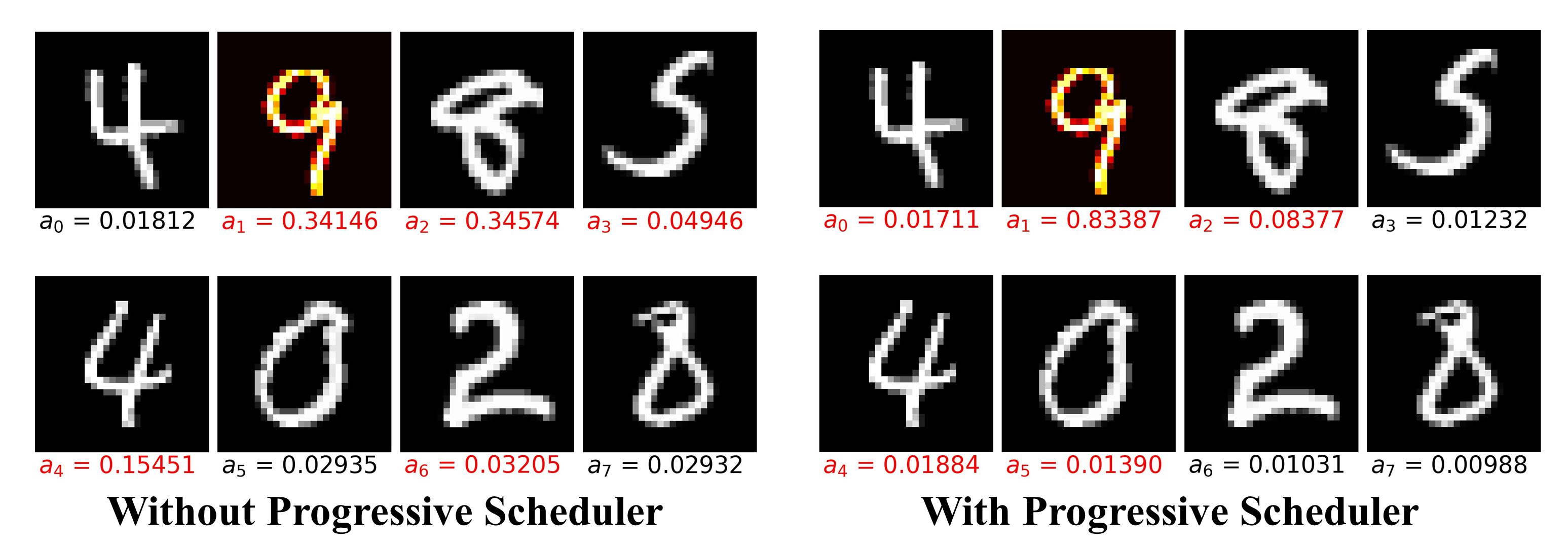} 
\caption{The attention localization of ABMIL.}
\label{fig:scheduler}
\vspace{-0.2cm}
\end{figure}
% Employing a fixed drop rate can misguide the localization of the MIL aggregator. The primary reason is that the MIL aggregator will identify negative instances as positive instances if instances with high attention weights are dropped out initially. 
Employing a fixed drop rate could misguide the localization of the MIL aggregator. The proposed progressive scheduler addressed this issue by allocating a gradually increasing drop rate during training. We experimented with MINIST, employing ABMIL as the aggregator method and comparing the fixed threshold and the progressive scheduler of PDL. As shown in Figure~\ref{fig:scheduler}, we visualized the attention map of ABMIL, and a fixed threshold method incorrectly identified `8' as a positive instance, even the attention weight beyond the target positive instance `9'. In the end, the PDL method with a fixed threshold achieved an accuracy of 0.87, while the PDL with a progressive scheduler attained an accuracy 0.96. The wrong positive instance recognition also impaired classification performance.

\section{Conclusion}
This study proposed a progressive dropout layer that may be integrated into prevalent MIL methods. Our investigation encompassed four datasets, achieving compelling results substantiating the effectiveness in mitigating overfitting and discovering latent features. Interestingly, we found that different MIL methods employed individually pre-trained ResNet weights. The diverse weight yielded significant disparities in experiment results, particularly in WSI. In the future, we will work on a unified paradigm for instance-level projectors: joint training with existing MIL aggregators. The new framework will enable the application of data augmentation on patches to solve overfitting.

\begin{appendices}
\section{Experiment Details}
We provide details of integrating PD into existing MIL methods, including ABMIL~\cite{ilse2018attention}, DTFD-MIL~\cite{zhang2022dtfd}, TransMIL~\cite{shao2021transmil}, DSMIL~\cite{li2021dual}.
\subsection{MIL Benchmark experiments setting}
\subsubsection{Embedding features} For the MUSK1 and MUSK2, A bag is constructed for each molecule, as instances. Each instance is represented with a 166-dimensional embedding feature. The FOX, TIGER, and ELEPHANT  datasets contain 200 bags of instance features. Each instance is associated with a 230-dimensional embedding feature.

\subsubsection{Detail of integrating PDL into existing MIL architectures} We employed three fully connected layers in the middle layer, each comprising 256, 128, and 64 hidden units. Each layer was equipped with a ReLU activation function and subsequently appended by a PDL layer. Following the passage of features through the intermediate layer, the ABMIL, ABMIL-Gate, and DSMIL aggregators processed the 64-dimensional embedding feature as their input. To compare original-based models, all aggregators adhere to the configuration outlined in paper~\cite{li2021dual}, with distinct embedding features of dimensions 230 and 166 directly employed as inputs. All experiments used the Adam optimizer with $2e^{-4}$ learning rates and $5e^{-3}$ weight decay and trained on cross-entropy loss for 40 epochs. The PDL parameters were identical to the paper description (section Implementation details).

\subsection{WSIs experimental setting}
\subsubsection{Preprocessing WSI}  To pre-process the WSIs datasets, every WSI is cropped into 224 × 224 patches without overlap to form a bag, in the magnifications of 20×. Background patches are discarded with threshold 30. 
\subsubsection{Embedding Network pretrained} The ResNet-18 was employed as the embedding network; it is worth noting that initially, we utilized pre-trained weights from ResNet50 trained on ImageNet following the paper~\cite{shao2021transmil}. However, due to the availability of various weight versions from PyTorch, some weights might be trainable in certain methods but not functional in others. As a result, we opted for ResNet-18, a set of weights that can be successfully trained across all MIL methods. In addition to the aforementioned pre-trained weights from ImageNet, we also incorporated weights obtained through a contrastive learning framework for self-supervised pre-training, SimCLR pre-trained. The self-supervised SimCLR manner employed the contrastive learning framework~\cite{simplecontrastivelearning} to pretrain the projector based on the training set, wherein contrastive loss training was implemented between extracted patches and corresponding two random data data-augmentation counterparts~\cite{li2021dual}. Each patch would project into a 512-dimensional embedding feature. 
\subsubsection{Detail of integrating PDL into existing MIL architectures}
We followed the parameter settings outlined in the original literature for the baseline experiments on the two WSI datasets, as shown in Table~\ref{tb:parameters}. For integrating PDL into these MIL methods, We employed three fully connected layers in the middle layer, each comprising 256, 128, and 64 hidden units. Each layer was equipped with a ReLU activation function and subsequently appended by a PDL layer. The input dimensions of these networks were adjusted to 64 while keeping the other parameters unchanged. All the experiments were trained on 200 epochs. The PDL parameters were identical to the paper description (section Implementation details). The experiments integrated with PDL also followed the parameter in Table~\ref{tb:parameters}.
\begin{table}[t]
\centering
\caption{Parameters setting for baseline MIL methods. Here employed the AdamW~\cite{adamW}, RAdam~\cite{radm}, CosineAnnealingLR~\cite{consin} to follow these MIL methods papers setting. }
\label{table2}
\resizebox{1\columnwidth}{!}{
\begin{tabular}{c|ccccc}
\toprule %
     & ABMIL & ABMIL-Gate & DSMIL & TransMIL & DTFD-MIL \\ \cmidrule (l ){1 -6}
    Optimizer & AdamW & AdamW &  AdamW & RAdam  & Adam \\ \cmidrule (l ){1 -6}
    Learning rate & $1e^{-4}$ & $1e^{-4}$&  $1e^{-4}$ & $1e^{-4}$ & $2e^{-4}$\\\cmidrule (l ){1 -6}
    Weight decay & $1e^{-4}$ & $1e^{-4}$ &  $5e^{-3}$ & $5e^{-3}$ &  $2e^{-3}$\\\cmidrule (l ){1 -6} 
    Optimizer scheduler & LookAhead & LookAhead &  CosineAnnealingLR & LookAhead  & MultiStepLR\\\cmidrule (l ){1 -6} 
    Loss function  & CrossEntropy & CrossEntropy &  CrossEntropy & CrossEntropy& CrossEntropy +  distille loss  \\
\bottomrule
\end{tabular}
}
\label{tb:parameters}
\end{table} 

\subsection{MNIST experiments setting}
\subsubsection{Eembedding Network} 
\begin{figure}[t]
\centering
\includegraphics[width=1\columnwidth]{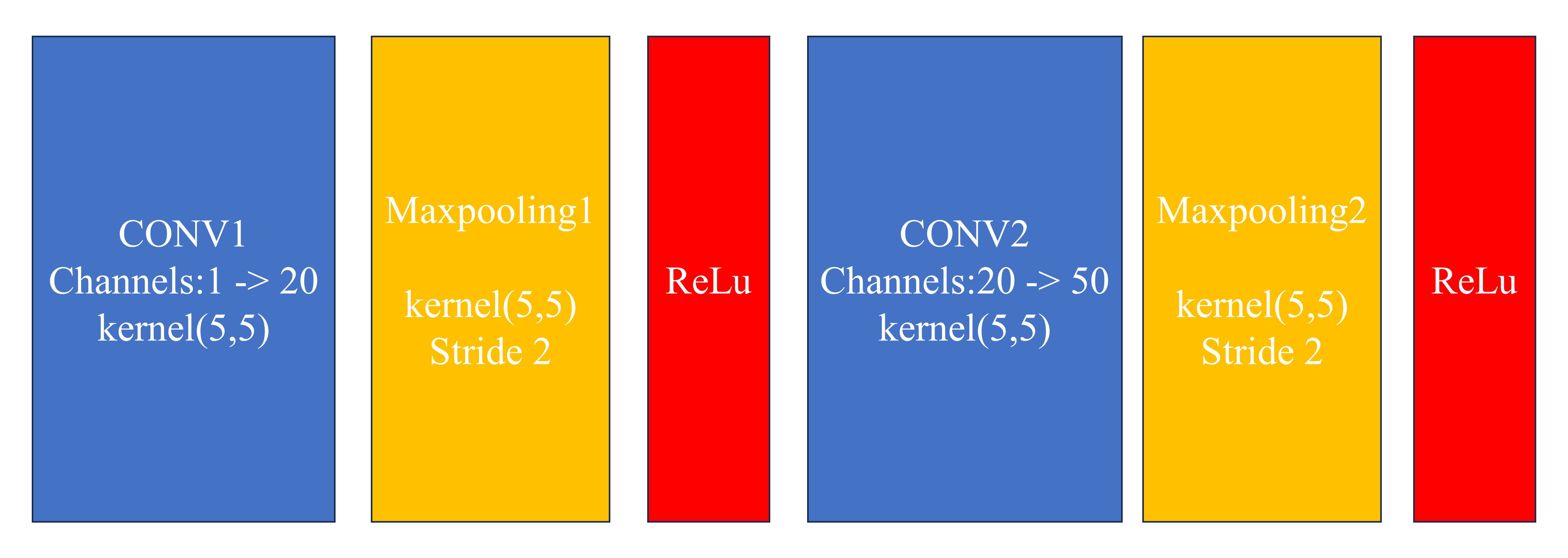} 
\caption{Emebedding network for MINIST experiment.}
\label{fig:mnist}
\vspace{-0.2cm}
\end{figure}

Each bag comprised a random assortment of 28 × 28 grayscale images extracted from the MNIST dataset. The count of images in a bag follows a Gaussian distribution, with the nearest integer value being considered. A bag receives a positive label if it contains one or more images with the label '9'. The average number of instances per bag was set to 20, with a standard deviation of 2. We employed embedding networks as those used in~\cite{ilse2018attention}(as shown in Figure~\ref{fig:mnist}). After through the embedding network, each instance was an 800-dimensional embedding feature. 

\subsubsection{Detail of integrating PDL into existing MIL architecture}
The experiment will be used in the Discussion section to analyze PDL rationality further. We employed two fully connected layers in the middle layer, each comprising 512 and 256 hidden units. Each layer was equipped with a ReLU activation function and subsequently appended by a PDL layer. We used the ABMIL aggregator as the primary research object. We used the ABMIL aggregator as the preliminary research object. It processed the 256-dimensional embedding feature as its input feature. The experiments used the Adam optimizer with $5e^{-4}$ learning rates and $1e^{-4}$ weight decay and trained on cross-entropy loss for 40 epochs. The PDL parameters were identical to the paper description (section Implementation details). Notably, owing to the constrained number of positive instances within the MNIST dataset, we have conducted experiments employing two PDL layers.

\section{Interpolation Methods}

% \subsection{Interpolation methods}
As mentioned in two components within our main body of the paper, the same interpolation method (Here is LOG) has been utilized to generate a set of drop rates for each instance and a set of global maximal drop rates for each epoch for "Dynamic assignment drop rate for each instance (DADR)" and "Progressive Learning Scheduler (PLS)." Here are three interpolation methods, as represented:
\begin{gather}
\begin{aligned}
 COS: &  P*(0.5*(1-cos(linspace(0, \pi , n)))) \\
 LOG: &  P/E* log_{G}(linspace(0, G^{E} - 1, n)+ 1) \\
 EXP: &  P/B *(G^{linspace(0, log_{G}(B+1), n)}  - 1)
\end{aligned}
\label{eqt:sample}
\raisetag{20pt}
\end{gather}
$linspace(min,max,num)$ returns $num$ evenly spaced samples from interval $[min, max]$, $n$ denotes the number of instances, and $E$, $G$, and $B$ are used to control the spacing towards of produced non-linear vector. Both three methods will generate a set of probability vector $\left \{ p_{1}, p_{2}, \dots, p_{n} \right \}$ from 0 to $P$. 
As shown in Figure~\ref{interpolation}, its main differences are used to generate a probability vector of different spacing at the beginning and end of the vector. 

\subsection{Details of experimental setting} 
\begin{figure}[t]
\centering
\includegraphics[width=1\columnwidth]{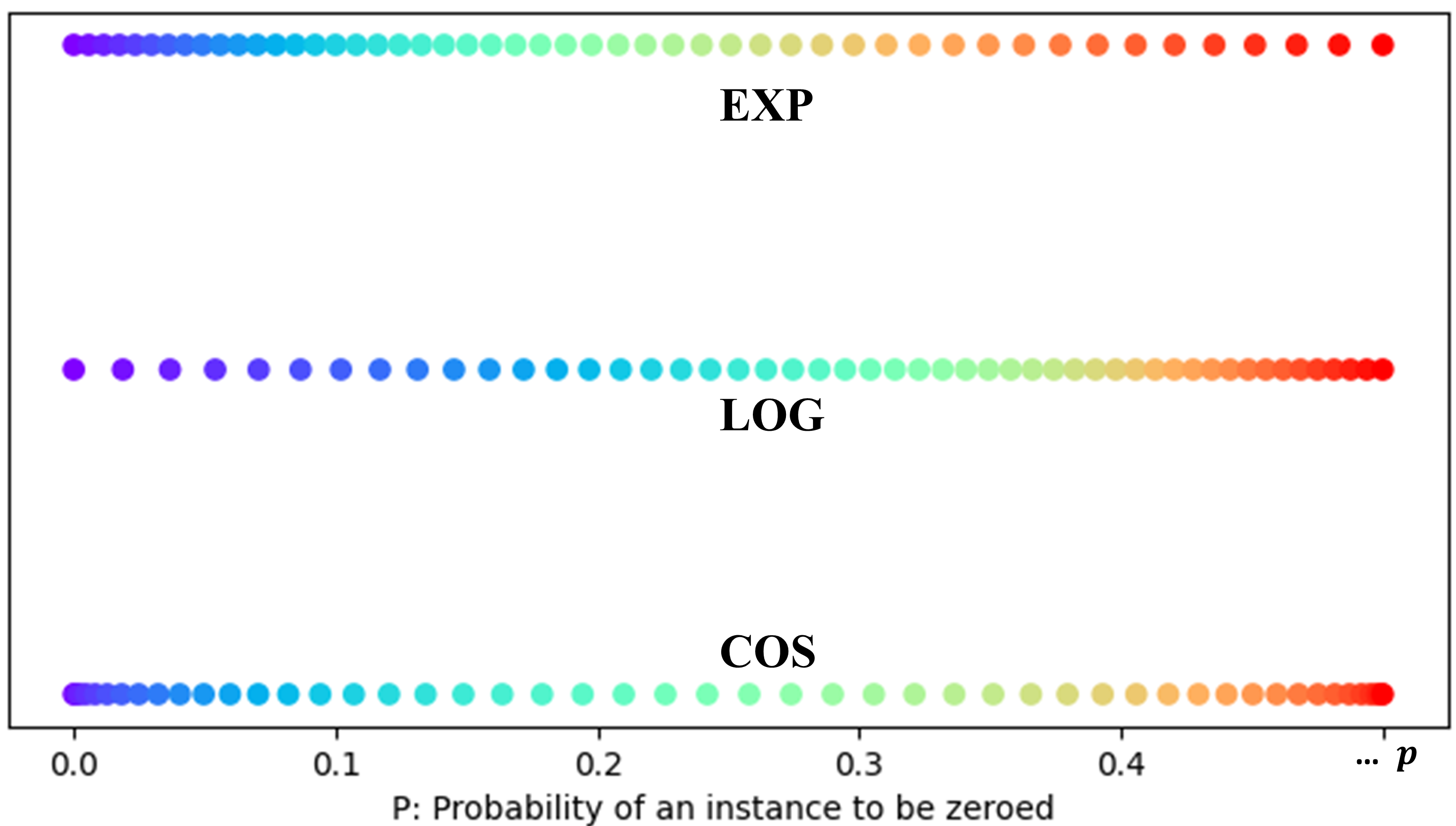}
\caption{Three interpolation methods}
\label{interpolation}
\end{figure}

In this experiments, we seted the 
$E$ = 0.5, $G$ = 10, and $B$ = 0.5 same with main body of paper. We employed the ABMIL as the MIL aggregator based on the CAMELOYON16 dataset. The pre-trained embedding network was Resnet-18 in SimCLR self-supervised manner. The Progressive learning scheduler adopted the three different interpolation methods to generate a set of $\left \{ P_{1}, P_{2}, \dots, P_{t} \right \}$ from a range of 0 to $P_{max} = 0.45$ for 200 epochs ($T = 200$), here we set the $P_{max}$ as $P$, $n = T$ in Equation~\ref{eqt:sample}. For the Dynamic drop rate assignment, we passed the $P_{t}$ for each epoch; the n is automatically adjusted based on the number of instances for each bag, which is obtained from the number of instances by the attention map, so here interpolation methods only required one parameter $P$ is enough in Equation~\ref{eqt:sample}.  Same with the WSIs experiment setting, we employed three fully connected layers in the middle layer, each comprising 256, 128, and 64 hidden units. Each layer was equipped with a ReLU activation function and subsequently appended by a PDL layer. The input size of ABMIL was adjusted to a 64-dimensional feature. All experiments employed the AdamW with a learning rate of $1e^{-4}$ and weight decay $1e^{-4}$ and adopted the Lookahead optimizer scheduler to adjust the learning rate. It was worth noting that three interpolation methods are satisfied with the progressive processing definition, as shown the Figure~\ref{interpolation}; they were a progressive increment function from 0 to P. 

\subsection{Experimental results}
\begin{table*}[htb]
\centering
\caption{Comparison results between three interpolation methods applied with Dynamic drop rate assignment(DADR) and Progressive Learning Scheduler(PLS) on the CAMELYON16 dataset.}
\label{table2}
\resizebox{1\linewidth}{!}{
\begin{tabular}{c|cc|cc|cc}
\toprule %
     & \multicolumn{2}{c}{DADR:LOG} & \multicolumn{2}{c}{DADR:COS}   & \multicolumn{2}{c}{DADR:EXP}\\
    \cmidrule(r){2-3} \cmidrule(r){4-5} \cmidrule(r){6-7}
     & Accuracy & AUC & Accuracy & AUC  & Accuracy & AUC \\ \cmidrule (l ){1 -7}
    PLS:LOG & 88.84 ± 0.43 & 91.09 ± 0.89  & 88.99 ± 1.00 & 90.85 ± 0.50 &  87.90 ± 1.60 & 90.62 ± 1.21\\
    PLS:COS &   87.75 ±  0.84 & 89.42 ± 0.54  & 88.21 ± 0.64 & 89.49 ± 1.36 & 88.06 ± 0.69 & 88.64 ± 0.77\\
    PLS:EXP & 87.28 ±  0.88 & 88.70 ± 0.86  & 88.06 ± 0.42 & 89.37 ± 1.49 & 87.13 ± 0.69 & 88.47 ± 0.53 \\

\bottomrule
\end{tabular}
}
\label{tb:exp}
\end{table*} 
To comprehensively evaluate the three interpolation methods, the DADR and PLS applied three interpolation methods (LOG, COS, EXP), respectively. We conducted five runs for each experiment and calculated the mean and variance. As depicted in table~\ref{tb:exp}, the three methods correspond to distinct interpolation approaches, each of which positions the generated points near 0, P, or concentrates them within an intermediate region. For datasets like WSI, the typical instance count ranges from 8000 to 10000, resulting in a relatively substantial number of positive instances. This substantial presence of positive instances is the underlying reason for the favorable performance of LOG with DADR. However, in comparison to the LOG approach, the EXP with PLS constrains the maximum global drop rate close to 0, leading to a scenario where only a small fraction of instances was dropped during most training epochs. Consequently, this significantly compromised its ability to mitigate overfitting. In relative terms, the performance of the COS with PLS remained relatively stable, although it fell short of the LOG with PLS efficacy. Here all interpolation methods could improve the performance compared with the baselines without PDL (\textbf{86.20 Accuracy and 87.52 AUC}). This performance variation was likely attributed to the specific datasets. We suggest selecting these three interpolation methods based on three main factors: i) the instance count of the dataset, ii) the proportion of positive instances, and iii) the number of training epochs.
\end{appendices}
\bibliographystyle{unsrt}  
\bibliography{references}

\end{document}